\documentclass{article}
\usepackage{spconf,amsmath,graphicx}

\usepackage{hyperref}
\usepackage{url}
\usepackage{xcolor}
\usepackage{booktabs}

\title{Overview of the ICASSP 2023 General Meeting Understanding and Generation Challenge (MUG)}
\name{
 \parbox{\linewidth}{\centering
Qinglin Zhang$^1$, Chong Deng$^1$, Jiaqing Liu$^1$, Hai Yu$^1$, Qian Chen$^1$ \\
Wen Wang$^1$, Zhijie Yan$^1$, Jinglin Liu$^2$, Yi Ren$^2$, Zhou Zhao$^2$}
}
\address{$^1$ Speech Lab of DAMO Academy, Alibaba Group $^2$ Zhejiang University\\
  $^1${\tt \{qinglin.zql, dengchong.d, mingzhai.ljq, yuhai.yu\}@alibaba-inc.com} \\
  $^1${\tt \{tanqing.cq, w.wang, zhijie.yzj\}@alibaba-inc.com} \\
  $^2$ {\tt \{jinglinliu, rayeren, zhaozhou\}@zju.edu.cn}}
\usepackage[pscoord]{eso-pic}
\newcommand{\placetextbox}[3]{
\setbox0=\hbox{#3}
\AddToShipoutPictureFG*{
\put(\LenToUnit{#1\paperwidth},\LenToUnit{#2\paperheight}){\vtop{{\null}\makebox[0pt][c]{#3}}}%
}%
}%

\begin{document}
\ninept
\maketitle
\begin{abstract}
ICASSP2023 General Meeting Understanding and Generation Challenge (MUG) focuses on prompting a wide range of spoken language processing (SLP) research on meeting transcripts, as SLP applications are critical to improve users' efficiency in grasping important information in meetings. MUG includes five tracks, including topic segmentation, topic-level and session-level extractive summarization, topic title generation,  keyphrase extraction, and action item detection. To facilitate MUG, we construct and release a large-scale meeting dataset, \emph{the AliMeeting4MUG Corpus}.
We review the dataset, track settings and baselines, and summarize the challenge results and major techniques used in the submissions. 
\end{abstract}

\placetextbox{0.5}{0.08}{\fbox{\parbox{\dimexpr\textwidth-2\fboxsep-2\fboxrule\relax}{\footnotesize \centering Accepted paper. \copyright  2023 IEEE. Personal use of this material is permitted. Permission from IEEE must be obtained for all other uses, in any current or future media, including reprinting/republishing this material for advertising or promotional purposes, creating new collective works, for resale or redistribution to servers or lists, or reuse of any copyrighted component of this work in other works.}}}

\vspace{-3mm}
\section{Introduction}
\label{sec:intro}
\vspace{-2mm}
Spoken language processing (SLP) applications
are crucial for distilling, organizing, and prioritizing information
and 
significantly improves users' efficiency in grasping important information in meetings
~\cite{MUG-data-paper}.
Meetings pose two great challenges to SLP. First, meeting transcripts exhibit \textbf{a wide variety of spoken language phenomena}, such as disfluencies, grammar errors,  
and incomplete/fragmented sentences due to speaker interactions, which causes drastic differences from written texts, the majority of training data of NLP models,  hence leads to dramatic performance degradation. Second, meeting transcripts are usually \textbf{long-form documents} with several thousand words or more, challenging to mainstay Transformer-based models.
Publicly available meeting corpora supporting SLP research are limited and on small scale, severely hindering advances of meeting SLP~\cite{MUG-data-paper}.
To fuel research on meeting SLP, we launch \textbf{ICASSP2023 General Meeting Understanding and Generation challenge (MUG)}\footnote{\url{https://modelscope.cn/headlines/article/52}}. To facilitate MUG, we construct and release the \textbf{AliMeeting4MUG Corpus (AMC)}.
To the best of our knowledge, AMC is the \textbf{largest meeting corpus in scale and facilitates the most SLP tasks}~\cite{MUG-data-paper}. MUG includes five tracks: Track1 Topic Segmentation (TS), Track2 Topic-level and Session-level Extractive Summarization (ES), Track3 Topic Title Generation (TTG),  Track4 Keyphrase Extraction (KPE), and Track5 Action Item Detection (AID).  We review the dataset, tracks and baselines, and summarize challenge results and major techniques used in submissions.

\vspace{-5mm}
\section{Dataset, Track Setting and Baselines}
\label{sec:corpus}
\vspace{-2mm}
\noindent \textbf{Dataset and Tracks}  Our paper~\cite{MUG-data-paper} describes the dataset and tracks for MUG in detail. AMC\footnote{\url{https://modelscope.cn/datasets/modelscope/Alimeeting4MUG/summary}} includes 654 collected Mandarin meetings with each meeting consisting of a 15- to 30-minute discussion by 2-4 participants covering diverse topics. 
The avg. session length is 10,772.5 tokens, showing the \emph{long-form document challenge}. The avg. \# turns per session is 376.3, indicating frequent speaker interactions. 
We create manual SLP annotations on manual transcripts of meeting recordings with manually inserted punctuation and manual speaker labels. Details of SLP annotations and analyses are in~\cite{MUG-data-paper}. 
 524 meetings are manually annotated for all 5 SLP tasks (TS,ES, TTG, KPE, AID) and the rest 130 meetings are manually annotated with only TS.  For Track2-5,  we partition the 524 meetings with all 5 SLP annotations into 295 sessions for \textbf{Train}, 65 sessions for \textbf{Dev}, 82 sessions for Stage1 test set (\textbf{exceptTS-Test1}) and 82 sessions as Stage2(Final) test set (\textbf{exceptTS-Test2}). For Track1, we use the same Train and Dev sets and partition the 130 meetings with only TS labels into 65 sessions as \textbf{TSonly-Test1} and 65 sessions as \textbf{TSonly-Test2}. 
\textbf{Track1-TS} requires segmenting the manual transcripts of a session into a sequence of non-overlapping, topically coherent segments. For evaluation, we use positive F$_1$, $P_k$, and \emph{WinDiff}(\emph{WD}). The leaderboard score is computed as $0.5\times\text{positive}F_1+0.25\times(1-P_k)+0.25\times(1-WD)$. \textbf{Track2-ES} requires extracting key sentences for each reference topic segment and the entire session. We report both average and best ROUGE-1,2,L scores based on the 3 references for topic- and session-level ES. The leaderboard score is the average of these 12 scores. \textbf{Track3-TTG} requires generating an informative and concise title for each reference topic segment. We report both average and best ROUGE-1,2,L scores based on the 3 references and the leaderboard score is the average of the 6 scores. \textbf{Track4-KPE} requires extracting top-K keyphrases from a session that can reflect its main content. We evaluate exact F$_1$ and partial F$_1$ at top-K ($K=10,15,20$) between predicted KPs and reference KPs, and the leaderboard score is the average of the 6 scores. \textbf{Track5-AID} requires detecting sentences containing information about actionable tasks. We report positive precision, recall, and F$_1$ and the leaderboard score is positive F$_1$. 

\begin{table}[ht]
\begin{center}
\scalebox{0.7}{
\begin{tabular}{l l l l}
\hline
\multicolumn{4}{l}{\textbf{Track 1 Topic Segmentation (TS)}} \\
Model  & positive F$_1$     & $1-p_k$   & 1-WD \\
PoNet & $23.2_{\pm0.51}$    & $0.589_{\pm0.012}$    & $0.569_{\pm0.019}$ \\
\hline
\hline
\multicolumn{4}{l}{\textbf{Track 2 Extractive Summarization (ES) (AVG)}}  \\
Model  & R-1 Avg./Best    & R-2 Avg./Best   & R-L Avg./Best \\
PoNet & $55.73_{\pm0.39}$/$63.87_{\pm0.62}$   & $33.66_{\pm0.55}$/$44.87_{\pm0.75}$  & $43.11_{\pm0.89}$/$54.82_{\pm0.96}$ \\
\hline
\multicolumn{4}{c}{\textbf{Topic-level ES}}  \\
Model  & R-1 Avg./Best    & R-2 Avg./Best   & R-L Avg./Best \\
PoNet &  $54.05_{\pm0.73}$/$66.60_{\pm0.91}$   & $36.87_{\pm0.84}$/$53.30_{\pm1.31}$   & $47.39_{\pm0.89}$/$63.06_{\pm1.15}$ \\
\hline
\multicolumn{4}{c}{\textbf{Session-level ES}}  \\
Model  & R-1 Avg./Best    & R-2 Avg./Best   & R-L Avg./Best \\
PoNet & $57.41_{\pm0.38}$/$61.14_{\pm0.72}$   & $30.45_{\pm0.48}$/$36.43_{\pm0.76}$   & $38.82_{\pm0.93}$/$46.57_{\pm1.08}$ \\
\hline
\hline
\multicolumn{4}{l}{\textbf{Track 3 Topic Title Generation (TTG)}}  \\
Model  & R-1 Avg./Best    & R-2 Avg./Best   & R-L Avg./Best \\
 PALM &  $28.44_{\pm0.40}$/$40.74_{\pm0.49}$   &  $15.26_{\pm0.31}$/$24.54_{\pm0.29}$  & $26.74_{\pm0.41}$/$39.04_{\pm0.50}$ \\
\hline
\hline
\multicolumn{4}{l}{\textbf{Track 4 Keyphrase Extraction (KPE)}}  \\
Model           & Exact/Partial F$_1$@10       & Exact/Partial F$_1$@15      & Exact/Partial F$_1$@20 \\
Structbert-NER            & $18.3_{\pm0.6}$/$32.0_{\pm0.4}$    & $23.2_{\pm0.8}$/$37.6_{\pm0.1}$   & $26.6_{\pm0.4}$/$41.2_{\pm0.3}$ \\
\hline
\hline
\multicolumn{4}{l}{\textbf{Track 5 Action Item Detection (AID)}}  \\
Model           &   positive P  & positive R    & positive F$_1$ \\
Structbert      &   $62.73_{\pm1.39}$           &   $71.1_{\pm0.35}$            &   $66.65_{\pm0.87}$                \\
\hline
\end{tabular}
}
\end{center}
\caption{\footnotesize{Baseline performance on the \emph{TSonly-Test1} set for Track1 and \emph{exceptTS-Test1} set for Track2-5. Each mean and std are computed on results from 5 runs with different random seeds.}}
\label{tab:baseline-results}
\end{table}

\noindent \textbf{Baseline Systems} We build baseline systems\footnote{\url{https://github.com/alibaba-damo-academy/SpokenNLP}} tackling the two key challenges of meeting SLP. We model TS and ES as sequence labeling tasks and AID as sentence classification task and compare BERT-base\footnote{\url{https://huggingface.co/bert-base-chinese}}, StructBERT-base~\cite{DBLP:conf/iclr/0225BYWXBPS20}, Longformer-base~\cite{DBLP:journals/corr/abs-2004-05150}\footnote{\url{https://huggingface.co/IDEA-CCNL/Erlangshen-Longformer-110M}} with linear complexity, and PoNet-base~\cite{DBLP:conf/iclr/TanCWZZL22}.  PoNet~\cite{DBLP:conf/iclr/TanCWZZL22} uses multi-granularity pooling and fusion 
for long sequence modeling and provides a linear-complexity drop-in replacement of self-attention.  PoNet achieves a good balance between complexity and transfer learning capability~\cite{DBLP:conf/iclr/TanCWZZL22}. StructBERT~\cite{DBLP:conf/iclr/0225BYWXBPS20} adds auxiliary pre-training objectives into BERT, which improve robustness to noisy word orders in spoken language. We find PoNet-based systems performs best on TS and ES and StructBERT-based system performs best on AID. Our TTG baseline uses the pre-trained PALM model~\cite{bi-etal-2020-palm}, which jointly pre-trains autoencoder and autoregressive language model hence better serves generation tasks. The KPE baseline uses StructBERT with CRF for sequence labeling, significantly outperforming the unsupervised YAKE.
Table~\ref{tab:baseline-results} reports the baseline results.

\vspace{-3mm}
\section{Summary of Track Results}
\label{sec:results}
\vspace{-2mm}
\noindent \textbf{Overall Results} 300+ developers participated in the MUG challenge and 47 teams submitted Stage2 (Final) evaluation submissions. 
Table~\ref{tab:official-results} reports leaderboard scores from Top-3 teams and our baselines on \emph{TSonly-Test2} for Track1 and \emph{exceptTS-Test2} sets for Track2-5.

\begin{table}[hbt]
\begin{center}
\scalebox{0.7}{
\begin{tabular}{@{}llllll@{}}
\toprule
Rank & \multicolumn{1}{c}{Track 1} & \multicolumn{1}{c}{Track2} & \multicolumn{1}{c}{Track3} & \multicolumn{1}{c}{Track4} & \multicolumn{1}{c}{Track5} \\ \midrule
1 & 48.84 & 52.56 & 33.77 & 45.07 & 63.89 \\
2 & 46.52 & 51.69 & 33.71 & 42.59 & 63.49 \\
3 & 42.92 & 49.58 & 31.98 & 38.43 & 61.31 \\
\hline
baseline & 41.01 & 49.57 & 27.50 & 41.48 & 61.15 \\ \bottomrule
\end{tabular}
}
\end{center}
\caption{\footnotesize{Results from Top-3 teams and our baselines on MUG Track1-5.}}
\label{tab:official-results}
\end{table}

\noindent \textbf{Tackle Two Key Challenges} We find the top performing teams all develop approaches to address the two key challenges of meeting SLP and these approaches yield most notable gains.  (1) \emph{To improve robustness to spoken language phenomena}, ES \#1 team~\cite{ES-top1} improves performance by exploring token-masking and span-masking with different masking rates for post-training DeBERTa. Many teams also improve performance on spoken language with preprocessing to remove disfluency and uninformative short sentences.
(2) \emph{To better model long-form documents}, TS \#1 team~\cite{TS-top1} achieves significant gains from modeling document-level context via inter-sentence Transformer on sentence representations over PoNet. TS \#2 team aggregates sentence representations in a paragraph and models inter-paragraph relations. ES \#1 team achieves superior performance through multi-task learning of topic- and session-level extractive summarization based on DeBERTa, as DeBERTa can handle input length up to 4096 tokens. ES \#2 team designs a hybrid PoNet+TransformerEncoder model, achieving a better balance of accuracy and efficiency over PoNet. 

\noindent \textbf{Explore Additional Data} (1) \emph{Data Augmentation}. TS \#1 team ensures consistent length distribution between their synthesized meeting data and AMC. TS \#2 team uses Maximum Mean Discrepancy Loss to ensure consistent distribution between AMC and their augmented data. (2) \emph{Explore Written Text Data}. TTG \#1 team~\cite{TTG-top1} develops multi-stage training to leverage knowledge from large models and additional data. They first pre-train the encoder-decoder CPT-large model on the news title generation data, then triple \emph{AMC Train} by using each of the 3 annotations as target and use the expanded data to fine-tune CPT-large from Stage1, and finally conduct joint fine-tuning and distillation with teacher as CPT-large from Stage2 and student as CPT-base pre-trained using the news title generation data. The Stage3 fine-tuning data is constructed on \emph{AMC Train} by selecting the title most similar to the other 2 titles as target. TTG \#2 team also first trains a pre-trained encoder-decoder model with two written text summarization corpora then fine-tunes with AMC.

\noindent \textbf{Other Techniques} (1) Many teams use focal loss to address \emph{imbalanced labeled data}. (2)  Many teams adopt adversarial training such as FGM to improve \emph{generalizaibility}. ES \#1 team improves generalizability using stochastic weight averaging on linear layers. (3) \emph{Task-specific}. KPE \#1 team~\cite{KPE-top1} jointly optimizes focal loss and regression loss (to fit model predicted scores to scores assigned by a W2NER module). Inspired by observations that most actionable items are acknowledged by other participants, 
AID \#1 team~\cite{AID-top1} 
expands model input with sentences and speaker labels from adjacent context. 

\vspace{-3mm}
\section{Conclusion}
\label{sec:conclusion}
\vspace{-2mm}
We provide an overview of ICASSP2023 General Meeting Understanding and Generation Challenge (MUG). We find approaches to address the two key challenges to meeting SLP yield most notable performance gains.
Exploring additional data, 
handling imbalanced labels, improving generalizability, and other task-specific approaches also contribute to performance gains.

\vspace{-2mm}
\scriptsize
\bibliographystyle{IEEEbib}
\bibliography{strings,mybib}

\end{document}